\documentclass{article}

    \PassOptionsToPackage{numbers, compress}{natbib}
     \usepackage[final]{nips_2018}

\usepackage{graphicx}
\usepackage{bbm}
\usepackage{amsmath}
\usepackage{caption}
\usepackage{subcaption}
\usepackage[utf8]{inputenc} 
\usepackage[T1]{fontenc}    
\usepackage{hyperref}       
\usepackage{url}            
\usepackage{booktabs}       
\usepackage{amsfonts}       
\usepackage{nicefrac}       
\usepackage{microtype}      
\usepackage{siunitx}

\title{Exploiting Data Hierarchy as a New Modality for Contrastive Learning}

\author{%
   Arjun Bhalla\thanks{These authors equally contributed to this work.}\\
   M.Eng. Computer Science \\
   \texttt{ab2383@cornell.edu} \\
   \And
   Daniel Levenson\footnotemark[1] \\
   M.S. Health Tech \\
   \texttt{dl999@cornell.edu} \\
   \And
   Jan Bernhard\footnotemark[1]\\
   M.Eng. Computer Science \\
   \texttt{jhb353@cornell.edu} \\
    \AND
    Anton Abilov\thanks{Dropped Deep Learning course mid-semester. Contributed to the brainstorming and data preprocessing.}\\
  M.S. Connective Media\\
  \texttt{aa2776@cornell.edu} \\
}

\begin{document}

\maketitle

\begin{figure}[hbt!]
    \centering
    \includegraphics[width=\textwidth]{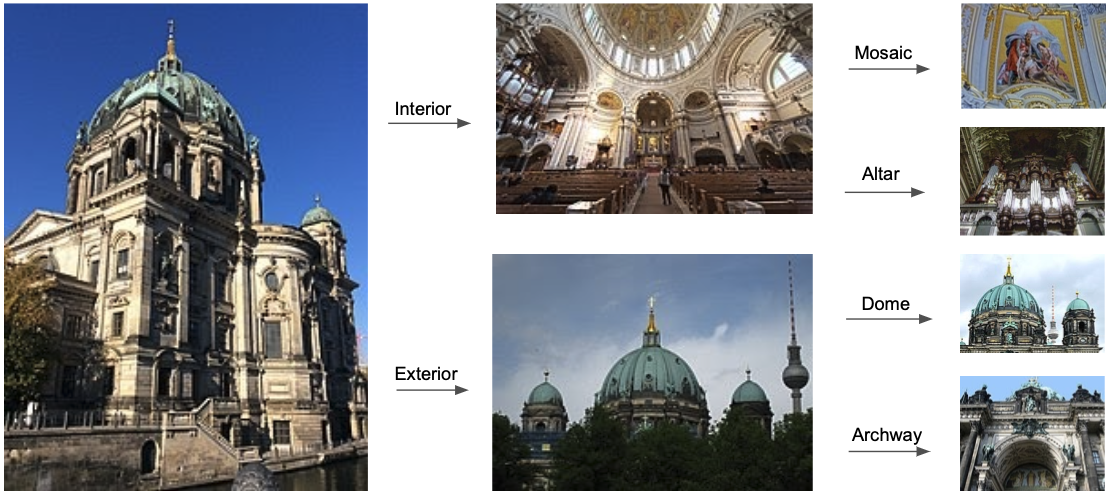}
    \caption{Images of the Berliner Dom used to demonstrate the hierarchical structure of the WikiScenes dataset.}
    \label{fig:banner}
\end{figure}
\begin{abstract}
 This work investigates how hierarchically structured data can help neural networks learn conceptual representations of cathedrals. The underlying WikiScenes dataset\footnote{We would like to thank \citet{wikiscenespaper} for granting us access to the WikiScenes dataset pre-release.} provides a spatially organized hierarchical structure of cathedral components. We propose a novel hierarchical contrastive training approach that leverages a triplet margin loss to represent the data's spatial hierarchy in the encoder's latent space. As such, the proposed approach investigates if the dataset structure provides valuable information for self-supervised learning. We apply t-SNE to visualize the resultant latent space and evaluate the proposed approach by comparing it with other dataset-specific contrastive learning methods using a common downstream classification task. The proposed method outperforms the comparable weakly-supervised and baseline methods. Our findings suggest that dataset structure is a valuable modality for weakly-supervised learning.
\end{abstract}

\section{Introduction}

An active open-question in Machine Learning research is how to learn useful representations from unstructured or loosely structured data. Data is often loosely structured in hierarchies: these data relate to each other, but don't have in a consistent structure. We investigate if there is value in including hierarchical structure as input into a weakly-supervised learning problem.

In order to perform this investigation, we leverage the novel WikiScenes dataset\cite{wikiscenespaper} which contains hierarchical representations of cathedrals, as illustrated by Figure \ref{fig:banner}. While cathedrals alone are a narrow concept, the dataset is expected to extend to other conceptual categories included on WikiMedia\footnote{\url{https://www.wikimedia.org}}. Moreover, we expect our contributions to apply to other conceptual categories, independently of WikiMedia. Our approach builds upon recent breakthroughs of learning multi-modal neural networks using contrastive learning \cite{radford2021learning, Zhang2021}. However, instead of using text as an additional modality for a contrastive loss, we relate images by using their relative location in the hierarchy. We introduce a training procedure that trains a neural network ordered by the data's hierarchical structure and demonstrate the utility of the learned latent space.

\section{Related Work}

In 2013, \citet{Bengio2013-jf} presented a list of strengths of representation learning. One characteristic of good representations is that they embed the hierarchical structure of the data, even from a weakly-supervised task. Additionally, good representations naturally form clusters in the learned latent space \cite{Bengio2013-jf}. Given our objective to learn characteristic representations of cathedrals from a hierarchical dataset, applying a representation learning approach is an obvious fit for the learning task. Yet, learning these representations is not trivial.

Contrastive learning, which is part of the representation learning landscape, has shown to learn powerful representations in a series of recent works \cite{He2020-jm, Chen2020-nc, Chen2020-tk, bachman2019learning, Henaff2019-hc, radford2021learning, Grill2020-io}. 
These approaches typically learn a latent representation of unlabeled data in which similar samples cluster together \cite{Falcon2020}. 
Many of them have only relied on a single modality \cite{He2020-jm, Chen2020-nc, Chen2020-tk, bachman2019learning, Henaff2019-hc, Grill2020-io} by augmenting samples to generate similar pairs for the learning process. Dropping the data augmentation component, \citet{radford2021learning}’s multimodal approach outperformed the previous start-of-the-art self-supervised contrastive model \cite{Chen2020-nc}, and even matched the performance of supervised methods on ImageNet. Recent work by Salesforce Research \cite{PCL} has focused on using hierarchical structural as input into a contrastive learning process by using an Expectation-Maximization objective to discover structure. For our approach, the hierarchy of the dataset provides an additional modality to inform the otherwise purely image-based self-supervised learning process. We apply an approach similar to the contrastive losses of weakly-supervised approaches and use the signal from the hierarchy as additional modality instead of relying on data augmentations.

Due to the incremental nature of our proposed hierarchical training procedure, we need to consider issue of catastrophic forgetting \cite{MCCLOSKEY1989109}. To address this issue, we experiment with revisiting the data of all layers repeatedly. This approach is similar to memory replay, which has been beneficial when considered to address catastrophic forgetting \cite{hayes2020remind}.

\section{Dataset}


\subsection{WikiScenes Dataset}

\begin{figure}[hbt!]
     \centering
     \begin{subfigure}[b]{0.66\textwidth}
         \centering
         \includegraphics[width=\textwidth]{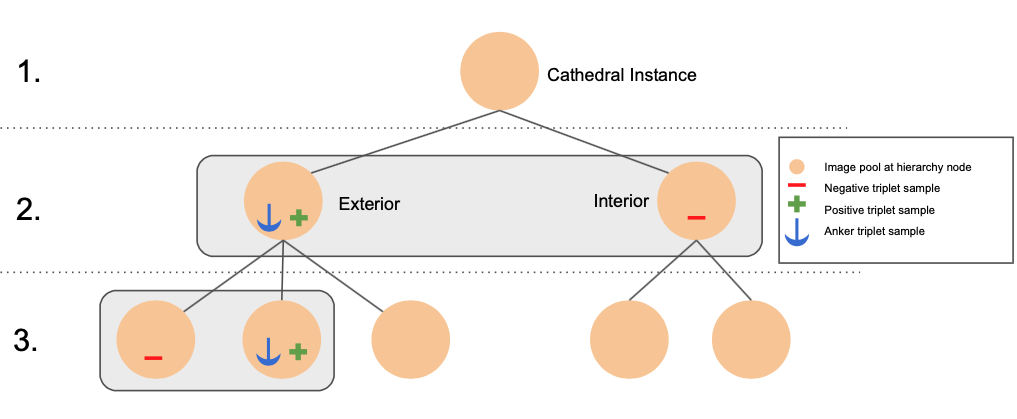}
         \caption{Dataset structure and contrastive sampling}
         \label{fig:data_hierarchy}
     \end{subfigure}
     \hfill
     \begin{subfigure}[b]{0.3\textwidth}
         \centering
         \includegraphics[width=\textwidth]{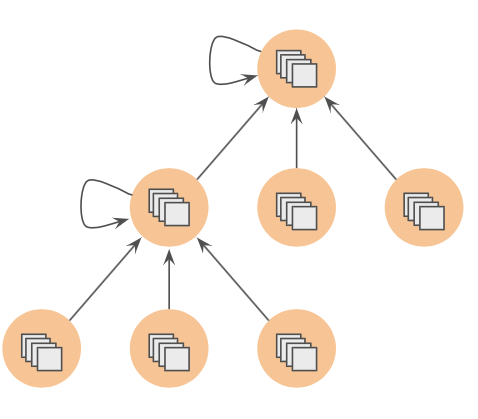}
         \caption{Images of descendent nodes contribute to their parent node's image pool.}
         \label{fig:pooling_logic}
     \end{subfigure}
     \hfill
     \caption{Visualizations of node structure underlying the hierarchical learning procedure.}
\end{figure}

The WikiScenes dataset consists of data from 99 cathedrals. Images can be associated with a leaf component of the hierarchy of the cathedral (e.g “Stained-glass windows in the Cathedral of St. John the Divine”) or with a parent component that itself is broken down into sub-components (e.g. “Interior of the Cathedral of St. John the Divine”). Figures \ref{fig:banner} and \ref{fig:data_hierarchy} illustrate the hierarchical structure of the dataset.
The hierarchy of information for each cathedral corresponds with the hierarchy of "concepts" encoded in Wikimedia\footnote{Example Wikimedia concept: "Views from Notre-Dame de Paris" - \url{https://commons.wikimedia.org/wiki/Category:Views_from_Notre-Dame_de_Paris}}. Notably, images are not mutually exclusive to a particular concept in the hierarchy: an image for the altar concept might appear in the images for the interior concept.
The WikiScenes dataset includes three data modalities: image, image description (text) and 3D image correspondence. Our work focuses on the image modality and the hierarchical modeling of the dataset.

\subsection{Contrastive Training Dataset Construction}\label{sec:Dataset_construction}

Elor et al.'s method focused on contrastive training using only examples that are labeled with a small set of "distilled concepts" \cite{wikiscenespaper}. Evaluating the impact of hierarchical training required a larger dataset that spans multiple hierarchy levels.
To explore the ability of our approach to generalize to other cathedrals, we consider datasets of multiple sizes: (1) Small: Only images of Notre-Dame ($\sim$23K images), (2) Medium: Images of Notre-Damn + six other cathedrals ($\sim$40K images), (3) Large: Images of all cathedrals ($\sim$104K images).
In order to maintain a rigorous evaluation approach, we construct datasets for this hierarchical contrastive pre-training which do not include any examples from the pre-defined image classification validation set.

\section{Method}

\subsection{Hierarchical Training Algorithm}

\subsubsection{Training Data Sampling}

We sample training data triplets by leveraging WikiScene's tree structure.
A node represents a directory in the WikiScenes dataset, e.g. Exterior, which holds a set of images directly associated to the concept represented by the dataset folder.
The edges between nodes represent folder and sub-folder relationships.
A node's image pool could be confined to only the images of its corresponding dataset directory and not any of its sub-directories. However, preliminary experiments showed that training on these image pools results in latent space collapse.
We mitigate this problem by creating a node's image pool from all images in its corresponding dataset directory and sub-directories, as shown in Figure \ref{fig:pooling_logic}, and revisiting a larger subset of images on consecutive hierarchy levels.

\subsubsection{Contrastive Loss} 

Our data sampling technique allows us to sample triplets of anchor, positive and negative samples given a node, its position in the dataset, and its image pool.
Given this triplet, we can train a neural network using a triplet margin loss \cite{10.5555/1577069.1577078}.
We sample images first by node and then by image to balance against nodes with an unbalanced distribution of images. As presented in Equation \ref{eqn:triplet_margin}, the triplet margin loss operates on three image samples: an anchor sample $(\mathbf{x}_a)$, a positive example, $(\mathbf{x}_p)$, from the same image pool as the anchor, and a negative example $(\mathbf{x}_n)$ from a node that shares the same parent as the anchor node. 
\begin{equation}
\label{eqn:triplet_margin}
    \mathcal{L}(\mathbf{x}_a, \mathbf{x}_p, \mathbf{x}_n; f) = max(0, \Vert f(\mathbf{x}_a )- f(\mathbf{x}_p)\Vert^2_2 - \Vert f(\mathbf{x}_a) - f(\mathbf{x}_n)\Vert^2_2 + \alpha)
\end{equation}
To compute the feature difference between the samples, the model, $f$, maps the input images onto three corresponding feature vectors. The margin hyperparameter $\alpha$ is tied to the current hierarchy level $h$, as shown in Equation \ref{eqn:margin_param}. 
\begin{equation}
\label{eqn:margin_param}
    \alpha(h) = (h_{max} - h)^2 + \alpha_{min} 
\end{equation}

\subsubsection{Level-Specific Triplet Margin}

The calculation of $\alpha$ requires the specification of $h_{max}$ and $\alpha_{min}$, which represent the deepest hierarchy level to consider and the minimum triplet margin at that deepest hierarchy level, respectively. By defining a level-specific triplet margin, each hierarchy level is evaluated with its corresponding unique loss function.

This approach aims to enforce large separation of high level concepts in their latent space representations and separate leaf concepts to a lesser extent: when $h$ is small compared to $h_{max}$, the difference is large and therefore the margin hyperparameter $\alpha$ is large, increasing loss and latent space separation on convergence. The goal of this approach is to learn strong representations by mapping the relationships between parent and child nodes onto a model's latent space.   
For the top level of the hierarchy, the model segments the latent space into three distinct classes: interior, exterior, and views. After stepping down the hierarchy once, the model segments each of these three concepts further into sub-classes, depending on the level and number of classes at that level. This pattern repeats until the training process reaches the predefined $h_{max}$ or concepts leaf nodes which do not have sub-classes. By reducing the triplet margin when moving from super- to sub-class, the sub-classes of a given super-class can separate while also maintaining a larger margin to the sub-classes of another super-class.

To keep the margin between super-classes after stepping to lower, sub-class hierarchy levels, it is important to occasionally revisit (replay) previous hierarchy layers. For that purpose, we introduced a replay regularization hyper-parameter $r_p \in [0.0, 1.0]$, which specifies the probability that a current batch is drawn from the current hierarchy level or any of the previous hierarchy levels. 

\subsection{Model Selection}

When exploring model architectures for the hierarchical training procedure, we focused on common image encoder models pretrained on ImageNet \cite{5206848}, such as ResNet50 \cite{he2015deep} and VGG-16 \cite{simonyan2015deep}. Preliminary experiments indicated that the fine-tuning of the convolutional layers resulted in poor performance. Consequently, we focus on the VGG-16, which allows us to freeze the convolutional layers and fine-tune the first and second fully connected layers of its classification module.

\subsection{Evaluation}

We evaluate the effectiveness of our novel hierarchical training approach in two ways: quantitatively via an image classification task and qualitatively via visualization of the latent space produced using the t-SNE \cite{JMLR:v9:vandermaaten08a} method.

\subsubsection{Quantitative Evaluation}

We evaluate the learned representations' quality by performing a supervised classification task on images from the WikiScenes dataset. To evaluate the learned representations, we freeze the pre-trained encoder networks, and train a single layer softmax classifier for the 10-class supervised single-class classification problem. We train the softmax layer with the same hyper-parameters for all models using the Adam optimizer \cite{kingma2017adam} and cross-entropy loss, batch size of 64, four epochs, and a learning rate of 0.001. Our baseline models for this evaluation are an ImageNet-pretrained VGG-16 model with no contrastive pre-training and a VGG-16 model with contrastive pre-training over just Wikiscenes leaf nodes. We directly compare our approach with \citet{wikiscenespaper}'s "Towers of Babel" encoder, which is a multi-modal contrastively trained ResNet-50 model \cite{he2015deep}. We then compare these baseline and comparison models against our best performing hierarchical-constrastively pretrained model.

For consistency with \citet{wikiscenespaper}, we report on the precision of the classification against the validation set averaged over classes, averaged over all examples, and broken out by each individual concept. The precision averaged over all classes, $mAP$, gives weight to concepts less represented in the validation set, and the precision averaged over all examples, $mAP^*$, gives every sample equal weight independent of its class.

\subsubsection{Qualitative Evaluation}

In order to investigate our hypothesis surrounding segmentation of the latent space, we perform a qualitative evaluation by using t-SNE \cite{JMLR:v9:vandermaaten08a} to reduce the feature maps from the encoder models to two dimensions and visualize them. This is done by first using Principal Component Analysis to reduce the dimensionality from 4096 to 50, and then further reduce this to two dimensions using t-SNE. We do this in order to provide a more numerically stable baseline on which to run t-SNE (as it gave worse results if reducing directly) and reduce noise. We then observe how different conceptual clusters form in the latent space. Thus, we are able to both qualitatively compare different models to explain the quantitative results, and use these visualisations to further understand what is happening during the training process.

\section{Results and Discussion}

This section presents the results of the various experiments. First, we present the downstream classification performance of our hierarchically trained, baseline, and comparison models. Second, we visualize t-SNE projections of the models' encoded features. Third, we compare the effect of hierarchy depth, dataset size, and replay frequency on the training outcome in an ablation study.

\subsection{Downstream Classification Task}

\begin{table}[hbt!]
    \centering
    \setlength\tabcolsep{0pt}
    \begin{tabular*}{\textwidth}{@{\extracolsep{\fill}} l*{20}{S[table-format=2.4]}}

    \toprule
     \small Model\hspace{0.2in} & \multicolumn{1}{c}{\small mAP} &
     \multicolumn{1}{c}{\small mAP$^*$} & \multicolumn{1}{c}{\small Facade} & \multicolumn{1}{c}{\small Window}& \multicolumn{1}{c}{\small Chapel}& \multicolumn{1}{c}{\small Organ}& \multicolumn{1}{c}{\small Nave}& \multicolumn{1}{c}{\small Tower}& \multicolumn{1}{c}{\small Choir}& \multicolumn{1}{c}{\small Portal}& \multicolumn{1}{c}{\small Altar}& \multicolumn{1}{c}{\small Statue}\\
     \midrule
        \small ImageNet VGG-16 & \small 38.9 \hspace{0.1in} & \small 47.7 \hspace{0.1in} & \small 53.3 & \small 71.4 & \small16.9 & \small\textbf{56.7} & \small\textbf{62.2} & \small8.7 & \small\textbf{56.4} & \small14.4 & \small 19.0 & \small 30.1\\
        \small Contrastive VGG-16 & \small41.0 & \small 49.1 & \small 83.5 & \small \textbf{94.7} & \small \textbf{20.7} & \small 46.8 & \small 51.5 & \small 10.1 & \small 42.9 & \small 13.9 & \small 14.8 & \small \textbf{30.5}\\
        \small Towers Of Babel \small \cite{wikiscenespaper} \hspace{0.1in} & \small35.6 & \small51.6 & \small\textbf{91.6} & \small72.9 & \small18.7 & \small 10.0 & \small 56.4 & \small 10.0 & \small 37.9 & \small 0 & \small \textbf{28.2} & \small 29.8 \\
        \small Ours & \small \textbf{41.6} & \small \textbf{51.9} & \small 88.5 & \small 93.6 & \small 17.8 & \small 46.1 & \small 48.5 & \small \textbf{11.9} & \small 45.0 & \small \textbf{15.6} & \small 25.1 & \small 24.3\\
        
    \end{tabular*}
    \vspace{0.25in}
    \caption{Classification Performance. The table above shows mean average precision (mAP), and per distilled concept average precision. mAP* is averaged over all images, not per concept. }
    \label{tab:classification_performance}
\end{table}

These results are comparable to WS-U results in \citet{wikiscenespaper}. Our best performing model is contrastively pretrained with a small set of cathedrals that are not within the validation set. While optimal models vary for each concept, our hierarchical model outperforms others on average. This implies a more consistent performance across classes.

Our optimal model is contrastively trained on the medium dataset described in Section \ref{sec:Dataset_construction} and leverages only the first level of the hierarchy for each cathedral. Our best model is trained with a replay regularization hyperparameter of 0.5, learning rate of 0.0001, and contrastive triplet batch size of 16.

We likely perform best overall because we benefit from using all the images associated with the provided cathedral instances during contrastive pre-training. The comparison model only exploits the data of the leaf nodes.

\subsection{t-SNE Latent Space Visualisations}

\begin{figure}[hbt!]
     \centering
     \begin{subfigure}[b]{0.32\textwidth}
         \centering
         \includegraphics[width=\textwidth]{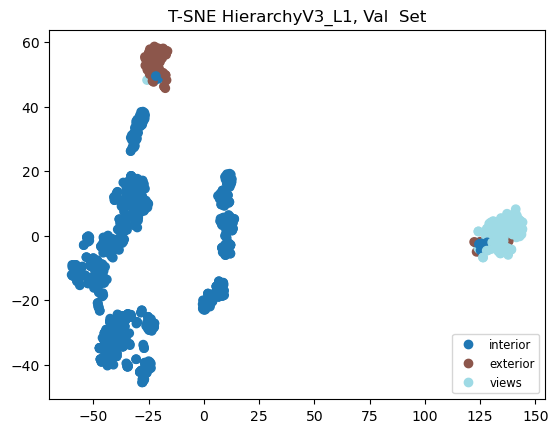}
         \caption{Our Method}
         \label{fig:hierarchy_l1}
     \end{subfigure}
     \hfill
     \begin{subfigure}[b]{0.32\textwidth}
         \centering
         \includegraphics[width=\textwidth]{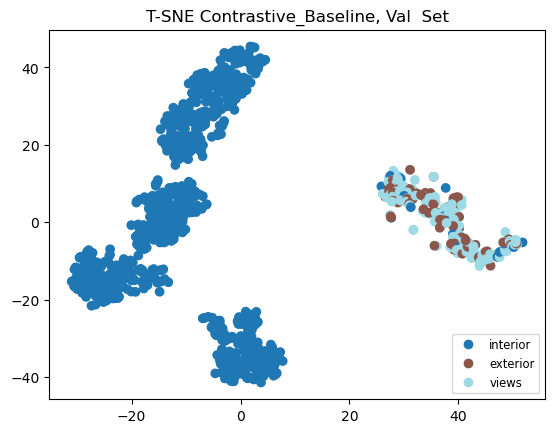}
         \caption{Contrastive VGG-16 Baseline}
         \label{fig:contrastive_baseline}
     \end{subfigure}
     \hfill
     \begin{subfigure}[b]{0.32\textwidth}
         \centering
         \includegraphics[width=\textwidth]{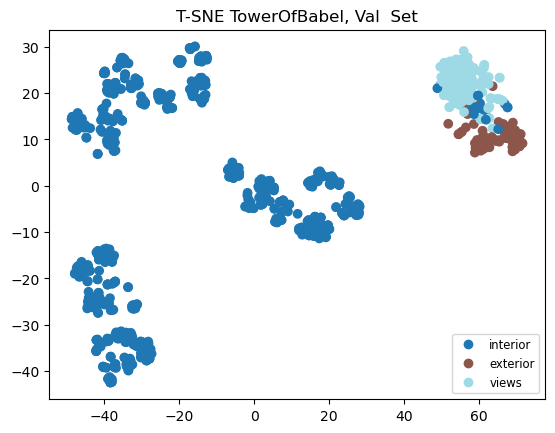}
         \caption{\citet{wikiscenespaper}'s Method}
         \label{fig:tower_l1}
     \end{subfigure}
        \caption{t-SNE plots of model feature maps for an unseen cathedral's first level}
        \label{fig:l1_featuremaps}
\end{figure}

As seen in Figure \ref{fig:hierarchy_l1}, our method provides clear feature-space separation between each of the classes provided by the hierarchy. The contrastive baseline (Figure \ref{fig:contrastive_baseline}) manages to separate the ``interior" from the other two classes fairly effectively, but mixes the ``exterior'' and ``views'' classes together. However, this is to be expected to a degree as they are similar in that they both include outdoor environments. Similarly, Figure \ref{fig:tower_l1} shows that \citet{wikiscenespaper}'s model generates significant separation between ``interior'' and other concepts, but maps ``exterior'' and ``views'' close together. Figure \ref{fig:l1_featuremaps} shows that our method can clearly and effectively distinguish between semantic classes. This implies that this hierarchical data structure modality can be useful for various applications and that our model is a good approach to exploiting this modality.

\begin{figure}[hbt!]
     \centering
     \begin{subfigure}[b]{0.32\textwidth}
         \centering
         \includegraphics[width=\textwidth]{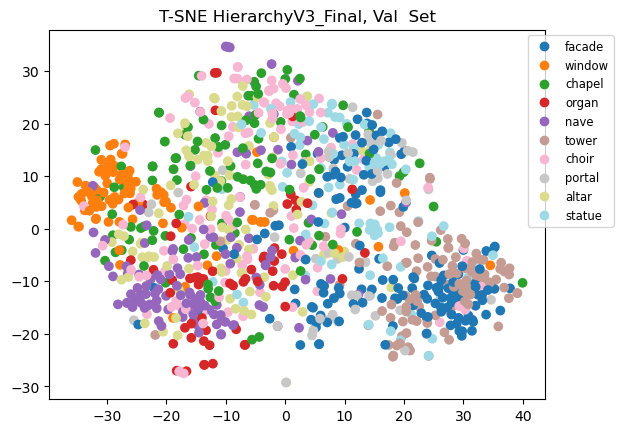}
         \caption{Our Method}
         \label{fig:hierarchy_classmap}
     \end{subfigure}
     \hfill
     \begin{subfigure}[b]{0.32\textwidth}
         \centering
         \includegraphics[width=\textwidth]{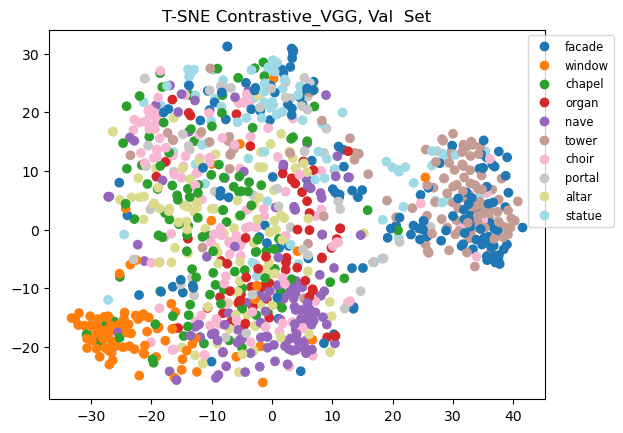}
         \caption{Naive Contrastive Baseline}
         \label{fig:contrastive_baseline_classmap}
     \end{subfigure}
     \hfill
     \begin{subfigure}[b]{0.32\textwidth}
         \centering
         \includegraphics[width=\textwidth]{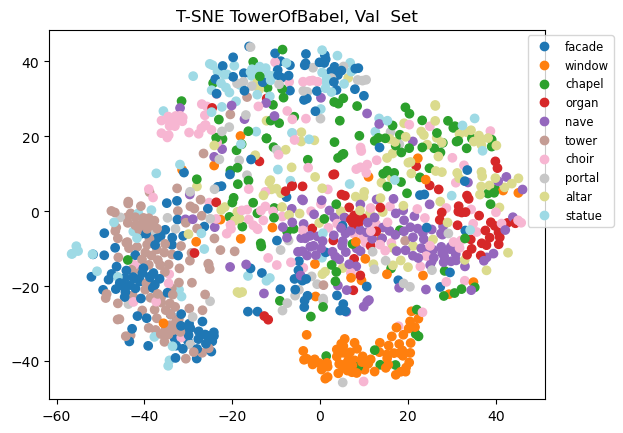}
         \caption{\citet{wikiscenespaper}'s method}
         \label{fig:tower_classmap}
     \end{subfigure}
        \caption{t-SNE plots of the latent space with respect to the downstream classification labels for an unseen cathedral.}
        \label{fig:l1_classmaps}
\end{figure}

As seen in Figure \ref{fig:hierarchy_classmap}, our method does not clearly have distinct clusters for each classified cathedral concept. However, upon further inspection, we see that some of the more prominent classes, such as facade, window, tower, and nave, are strongly concentrated in distinct regions in feature space. There are also weaker clusters for other concepts. We see a similar occurrence in Figure \ref{fig:contrastive_baseline_classmap}, for the contrastive baseline, but with weaker clusters for concepts where our model is strong, such as window and nave. \citet{wikiscenespaper}'s method clearly has the strongest segmentation, but even this suffers from a fair amount of class overlap and ambiguity.

Furthermore, this implies that despite not explicitly training on these classes, our method can still learn conceptual relationships effectively. In turn, this suggests that learning from the data's structural hierarchy modality using our method has strong conceptual representational ability. It also further supports our hypothesis that gradually segmenting the latent space with our method groups spatially similar concepts despite being given no information about their semantic relationship (supervised classification labels). However, when comparing these plots, it is important to note that \citet{wikiscenespaper}'s model is trained with the downstream classification labels as part of the contrastive loss function, whereas our model has no knowledge of these classes, which we believe is a major reason that they have more robust latent space separation.

\subsection{Ablation Study}

\begin{figure}[hbt!]
     \centering
     \begin{subfigure}[b]{0.32\textwidth}
         \centering
         \includegraphics[width=\textwidth]{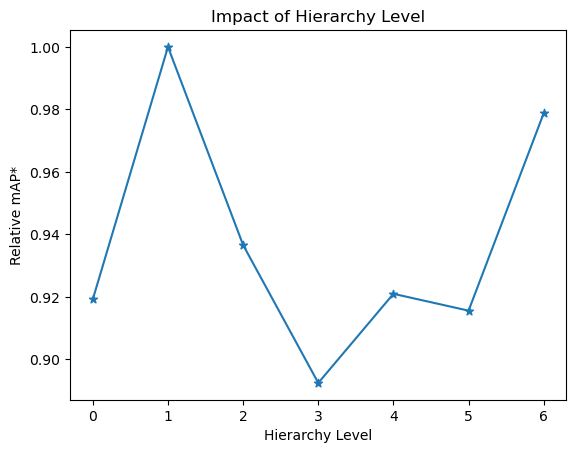}
         \caption{}
         \label{fig:dataset_size}
     \end{subfigure}
     \hfill
     \begin{subfigure}[b]{0.32\textwidth}
         \centering
         \includegraphics[width=\textwidth]{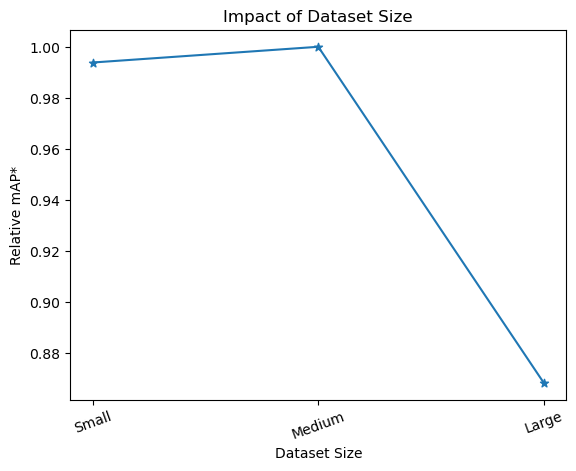}
         \caption{}
         \label{fig:hierarchy_level}
     \end{subfigure}
     \hfill
     \begin{subfigure}[b]{0.32\textwidth}
         \centering
         \includegraphics[width=\textwidth]{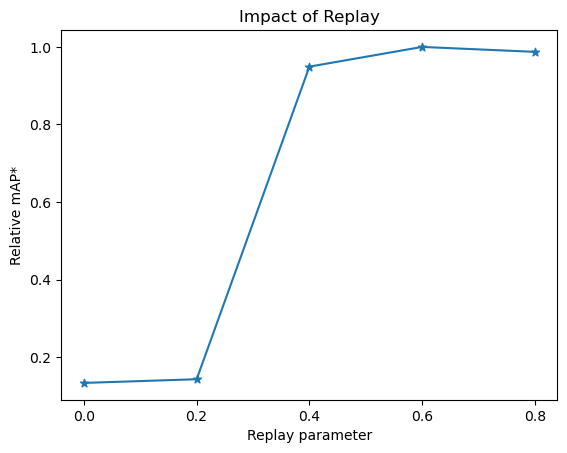}
         \caption{}
         \label{fig:replay}
     \end{subfigure}
        \caption{Each star marker represents a trained encoder model evaluated on the classification task. Relative $mAP*$ reflects the relative regression compared to the best classification performance of all models in the plot. For example, a Relative $mAP*$ of 0.10 is 90\% worse than the best performing model.}
        \label{fig:ablation_study}
\end{figure}

Figure \ref{fig:ablation_study} presents the impact of hierarchical information, dataset size, and replay frequency on the downstream classification performance. The star markers in each plot represent one full training run including contrastive pre-training and linear probing to assess downstream classification performance. Within a given plot, only the specified parameter changed between training runs. 
The y-axes of Figure \ref{fig:ablation_study} specify classification performance relative to the top performing model in the set presented a given plot. 
Figure \ref{fig:hierarchy_level} shows the impact of hierachical training on the downstream task. Hierarchy level zero corresponds to no hierarchical pre-training, and, thus, represents the vanilla ImageNet VGG-16. A significant finding is that training a model on just the first and a model on all hierarchical layers result in significant performance improvements, whereas training on layers two through five result in comparable performances to the vanilla model. This indicates that the concepts of the intermediate hierarchy levels provide less strong of a signal. This observation likely relates to the inconsistency in the number of hierarchy levels per cathedral instance, which is discussed in the next paragraph.

The impact study of dataset size suggests that our model is performing best for the medium sized dataset, as illustrated by Figure \ref{fig:dataset_size}. This regression in performance can be explained with the varying number of hierarchy levels between top level concepts, such as ``interior", and leaf node concepts, such as ``altar.'' The large dataset includes many cathedral instances for which the hierarchical structure is flat which means leaf node concepts may occur as early as the second hierarchy level. Consequently, some leaf node concepts are evaluated with a large margin loss, which likely causes deterioration of performance. To address this shortcoming of our method, we would need to introduce an adaptive hierarchy logic that sets $h_{max}$ programmatically based on the depth of the currently considered cathedral instance. The medium sized dataset is likely the best performing because it comprises cathedral instances of similar hierarchy depth. Consequently, the performance gains due to the additional data outweigh the performance regressions due to inconsistent hierarchy depths. 

Including replay into the training procedure resulted in significant improvements on downstream classification performance. Figure \ref{fig:replay} shows that a replay probability above 0.4 improves training outcomes significantly. A replay probability of 0.0 indicates that previous hierarchy levels are never revisited. The impact of the replay regularization parameter indicates that the current approach is prone to catastrophic forgetting. The given replay approach seems to mitigate this failure mode. 

\section{Limitations}
Our proposed hierarchical training method suffers from multiple limitations. First, its downstream task performance doesn't scale with dataset size, as shown in the ablation study. Second, we only explore this training approach on image data. It is unlikely that this exact method can be successfully applied to other modalities, such as language. Second, our method is specific to the spatial hierarchy of the WikiScenes dataset. To evaluate our method's full merit, it is necessary to look at other hierarchical datasets, e.g., the ImageNet database \cite{5206848}. Using ImageNet, one could investigate if hierarchical training would work for semantic hierarchical structures of animal breeds, such as dogs. Another limitation of this work is that hierarchical training could not successfully fine-tune or train from scratch convolutional layers and should be explored further.
\section{Future Work}
Given these promising initial results, there are many avenues for future work. The next step in our research would be to apply this to other structured datasets as more inherently structured WikiMedia datasets with different content is released. Along this same vein, it would be interesting to both run ImageNet classification on our method as it stands, and apply this method to ImageNet's purely semantic hierarchy, to see if our hypothesis for spatial hierarchy reasoning applies to other types of hierarchies. To improve latent feature representations, it is important to attempt more computationally intensive work on this matter - for example, training a model purely from scratch instead of using an ImageNet pretrained initialisation, and conducting more thorough hyperparameter sweeps to explore the effect of a wider spectrum of hyperparameters on the latent space. For example, we anticipate that different margin distance metrics and hierarchy scheduling functions might have a positive impact on the overall model performance.

\section{Conclusion}

We proposed a new training method that uses the hierarchical structure of data scraped from WikiMedia as a modality to inform weak supervision. The experiment's results suggest that considering hierarchical training can improve performance on downstream tasks. Future work is required to assess the full merit of this approach and address some issues around downstream task performance regressions for hierarchical training on large datasets. Nevertheless, the proposed method performs slightly better than comparable contrastive learning methods. Consequently, if given a structured dataset, hierarchical training is a competitive alternative to other self-supervised pre-training methods.
\newpage
{\small
\bibliography{bibliography}
}
\end{document}